\documentclass{article}

\usepackage[utf8]{inputenc} 
\usepackage[T1]{fontenc}    
\usepackage{hyperref}       
\usepackage{url}            
\usepackage{booktabs}       
\usepackage{amsfonts}       
\usepackage{nicefrac}       
\usepackage{microtype}      
\usepackage{amsthm}
\usepackage{amsmath}
\usepackage{graphicx}
\usepackage{pifont}
\usepackage{multirow}
\usepackage{multicol}
\usepackage{paralist}

\usepackage{arxiv}

\DeclareMathOperator*{\kthmin}{kthmin}
\DeclareMathOperator*{\kthmax}{kthmax}

\usepackage{amsmath,amsfonts,bm}

\def\ceil#1{\lceil #1 \rceil}

\def\1{\bm{1}}

\def\vzero{{\bm{0}}}

\def\va{{\bm{a}}}
\def\vb{{\bm{b}}}

\def\vg{{\bm{g}}}

\def\vx{{\bm{x}}}

\def\vz{{\bm{z}}}
\def\vdelta{{\bm{\delta}}}
\def\vlambda{{\bm{\lambda}}}

\DeclareMathAlphabet{\mathsfit}{\encodingdefault}{\sfdefault}{m}{sl}
\SetMathAlphabet{\mathsfit}{bold}{\encodingdefault}{\sfdefault}{bx}{n}

\newcommand{\R}{\mathbb{R}}

\usepackage{color}
\usepackage[linesnumbered,algoruled,boxed,lined]{algorithm2e}

\usepackage{subcaption}

\newtheorem{lemma}{Lemma}

\title{Evaluating the Robustness of Nearest Neighbor Classifiers: A Primal-Dual Perspective}

\author{
Lu Wang \\
Nanjing University \\
\texttt{wangl@lamda.nju.edu.cn} \\
\And
Xuanqing Liu \\
University of California, Los Angeles \\
\texttt{xqliu@cs.ucla.edu} \\
\And
Jinfeng Yi \\
JD AI Research \\
\texttt{yijinfeng@jd.com} \\
\And
Zhi-Hua Zhou \\
Nanjing University \\
\texttt{zhouzh@lamda.nju.edu.cn} \\
\And
Cho-Jui Hsieh \\
University of California, Los Angeles \\
\texttt{chohsieh@cs.ucla.edu}
}

\begin{document}

\maketitle

\begin{abstract}
We study the problem of computing the minimum adversarial perturbation of the Nearest Neighbor (NN) classifiers. Previous attempts either conduct attacks on continuous approximations of NN models or search for the perturbation by some heuristic methods.
In this paper, we propose the first algorithm that is able to compute the minimum adversarial perturbation. 
The main idea is to formulate the problem as a list of convex quadratic programming (QP) problems that can be efficiently solved by the proposed algorithms for 1-NN models. Furthermore, we show that dual solutions for these QP problems could give us a valid lower bound of the adversarial perturbation that can be used for formal robustness verification, giving us a nice view of attack/verification for NN models. For $K$-NN models with larger $K$, we show that the same formulation can help us efficiently compute the upper and lower bounds of the minimum adversarial perturbation, which can be used for attack and verification. 


\end{abstract}

\section{Introduction}
Adversarial robustness of neural networks has been extensively studied in the past few years. Given a data point, adversarial attacks are developed to construct small imperceptible input perturbations to alter the predicted label~\cite{szegedy2013intriguing,goodfellow2015explaining,carlini2017towards,athalye2018obfuscated,biggio2018wild}. On the other hand, robustness verification algorithms are also developed to compute a ``safe region'' around the point such that the prediction is provably unchanged within such region~\cite{kolter2017provable,weng2018towards,zhang2018efficient,Dvijotham18}. An attack algorithm can be viewed as finding an upper bound of the ``minimum adversarial perturbation'' while a verification algorithm finds a lower bound of this value.
In fact, robustness verification is often more important than attacks, since a verifiable behavior is required for mission-critical systems. 
For neural network models, due to non-convexity, both attack and verification cannot reach the minimum adversarial perturbation, and there is still a huge gap between the (computable) upper and lower bounds~\cite{salman2019convex}.

We study the problem of evaluating the robustness of the Nearest Neighbor (NN) classifiers. As a non-continuous step function, NN classifiers are very different from neural networks, and the neural network attack and verification methods cannot be directly applied to them. Previous attempts on attacking nearest neighbor models either use some simple heuristics~\cite{sitwarin2019robustness} or apply gradient-based attacks to some continuous substitute models of NN~\cite{sitwarin2019robustness,PapernotMG16,dubey2019defense}. 
Unfortunately, these attacks are far from optimal and do not have any theoretical guarantee. To the best of our knowledge, there is no existing approach on computing the minimum adversarial perturbation that can change an NN classifier's output, and there is no existing verification method that can compute a meaningful lower bound of the safe region. 

In this paper, we first study the $1$-NN classifier and show that finding the minimum adversarial perturbation can be formulated as a set of convex quadratic programming (QP) problems, and a solution can be computed in {\bf polynomial time}~\cite{kozlov1980polynomial}. This is quite different from neural networks or tree-based models where finding the minimum perturbation has shown to be NP-hard~\cite{katz2017reluplex,kantchelian2016evasion}. Furthermore, our formulation provides a very clean view of attack and verification for nearest neighbor classifiers. An attacker could solve any QP problem and any feasible solution will be a successful attack; a verifier could solve the dual of these problems and any feasible solution set will lead to a guaranteed lower bound of the minimum adversarial perturbation. 
Moreover, the primal minimum and dual maximum will match at the value of the minimum adversarial perturbation. We show that the QP problems can be solved efficiently by greedy coordinate ascent, and based on this primal-dual perspective,  we further provide several screening rules to speed up the quadratic solvers. 

When extending to $K$-NN models with $K>1$, our QP formulation will have the number of constraints growing exponentially with $K$.
However, we can still approximately solve the primal problems, and that will give an attack algorithm outperforming previous works.  Furthermore, we propose a way to set dual feasible solutions to provide a tight lower bound of the minimum adversarial perturbation without solving any problem. This leads to an efficient $K$-NN verification algorithm that works for any $K$. 


We conduct experiments on real datasets and have the following interesting findings: 
\begin{compactitem}
    \item For $1$-NN models, our proposed algorithm can efficiently compute the minimum adversarial perturbation. Our algorithm is provably optimal, achieves much smaller value, and is more efficient than previous attack methods. Also, this is the first robustness verification method for NN models. 
    \item For $K$-NN models with larger $K$, computing the exact minimum adversarial perturbation is still challenging, but our formulation provides an efficient attack algorithm, which outperforms previous attack methods. More importantly, our dual problems lead to an efficient verification algorithm to compute the lower bound of adversarial perturbation and have time complexity independent to $K$. Experiments show that the bounds are reasonably tight. 
   \item Equipped with our algorithm, we accurately compute the robust error bound of the 1-NN model on MNIST and Fashion-MNIST. We find that a simple 1-NN model can achieve better robust error than CNN on these data. 
\end{compactitem}

\section{Related work}

\paragraph{Adversarial robustness of neural networks. }
Adversarial robustness of neural networks has been studied extensively in the past few years. 
To evaluate the robustness of neural networks, attack algorithms are developed to find adversarial examples that are close to the original example~\cite{carlini2017towards,goodfellow2015explaining,madry2017towards,chen2017zoo,cheng2019query}. However, due to the non-convexity of neural networks, these attacks cannot reach the minimum perturbation so they can only provide some upper bound of robustness and cannot provide any robustness guarantee. For safety-critical applications such as real-world control systems, it is essential to have robustness guarantees such that we know the prediction is provably unchanged within a certain distance. This motivates recent studies on neural network verification which aims to compute a lower bound of the minimum adversarial perturbation~\cite{weng2018evaluate,kolter2017provable,weng2018towards,gehr2018ai2,wang2018efficient,zhang2018efficient,zhang2019recurjac}. Also, many of these robustness verification bounds can be incorporated in the training procedure to obtain  ``verifiable'' networks~\cite{kolter2017provable,wong2018scaling,mirman2018differentiable}. 


\paragraph{Adversarial robustness of nearest neighbor classifiers. }
Adversarial robustness of nearest neighbor classifiers is less studied.
Unfortunately, the algorithms mentioned above designed for neural networks cannot be directly applied to NN models since NN models are discrete step functions.  
\cite{wang2018analyzing} discussed the robustness of $K$-NN from the theoretical perspective and showed that the robustness of $K$-NN can approach the Bayesian optimal classifier. 
To compute an upper bound of the minimum adversarial perturbation (or equivalently, attack), 
\cite{PapernotMG16} proposed to employ a differentiable substitute for attacking 1-NN models; 
\cite{sitwarin2019robustness} proposed some heuristic methods and another gradient-based model to attack another kind of continuous substitute of $K$-NN. 
We will show in Section \ref{sec:background} that they cannot obtain the minimum adversarial perturbation and in experiments that they lead to loose upper bounds. 
On the other hand, to the best of our knowledge, there is no existing approach on computing the minimum adversarial perturbation or its lower bound, so our work is the first to verify the NN models. Finally, there are some recent work using NN models for defense, including~\cite{papernot2018deep,dubey2019defense}. However, they usually combine NN with neural network models, which are out of the scope of this paper. 

\section{Background and motivation}
\label{sec:background}
First, we set up notations for the Nearest Neighbor (NN) classifiers. Assume there are $C$ labels in total. 
We use $\{(\vx_1, y_1), \dots, (\vx_n, y_n)\}$ to denote the database where each $\vx_i$ is a $d$-dimensional vector and $y_i\in \{1, \dots, C\}$ is the corresponding label. A $K$-NN classifier 
$f: \R^d \rightarrow \{1, \dots, C\}$ maps a test instance to a predicted label. Given a test instance $\vz\in \R^d$, the classifier will  first identify the $K$-nearest neighbors $\{\vx_{\pi(1)}, \dots, \vx_{\pi(K)}\} $ based on the Euclidean distance $\|\vx_i - \vz\|$ and then predict the final label by  majority voting among $\{y_{\pi(1)}, \dots, y_{\pi(K)}\}$.

Next we define the notions of adversarial robustness, attack, and verification. Given a test sample $\vz$ and without loss of generality, we assume it is correctly classified as class-$1$ by the NN model. An adversarial perturbation is defined as $\vdelta\in \R^d$ such that $f(\vz + \vdelta) \neq 1$. An attack algorithm aims to find the minimum-norm adversarial perturbation, and its norm is
\begin{equation}
\epsilon^* = \big\{ \min_{\vdelta} \|\vdelta\| \ \text{ s.t. } \ f(\vz+\vdelta) \neq 1 \big\}.
\label{eq:attack}
\end{equation}
A verification algorithm aims to find a lower bound $r$ such that 
\begin{equation*}
    f(\vz+ \vdelta) = 1,  \ \ \ \ \forall \|\vdelta\|\leq r.
\end{equation*}
Clearly, by definition the maximum lower bound $r^*$ will match with the minimum perturbation norm $\epsilon^*$ if we have optimal attack and verification. 
We will mainly focus on $\ell_2$ norm but will briefly talk about how to extend to $\ell_\infty$ and $\ell_1$ norms later. Also, we will focus on $1$-NN first and then generalize to $K>1$ later. 

\paragraph{Failure cases of previous attack methods.}


\begin{figure}
    \centering
    \includegraphics[width=300pt]{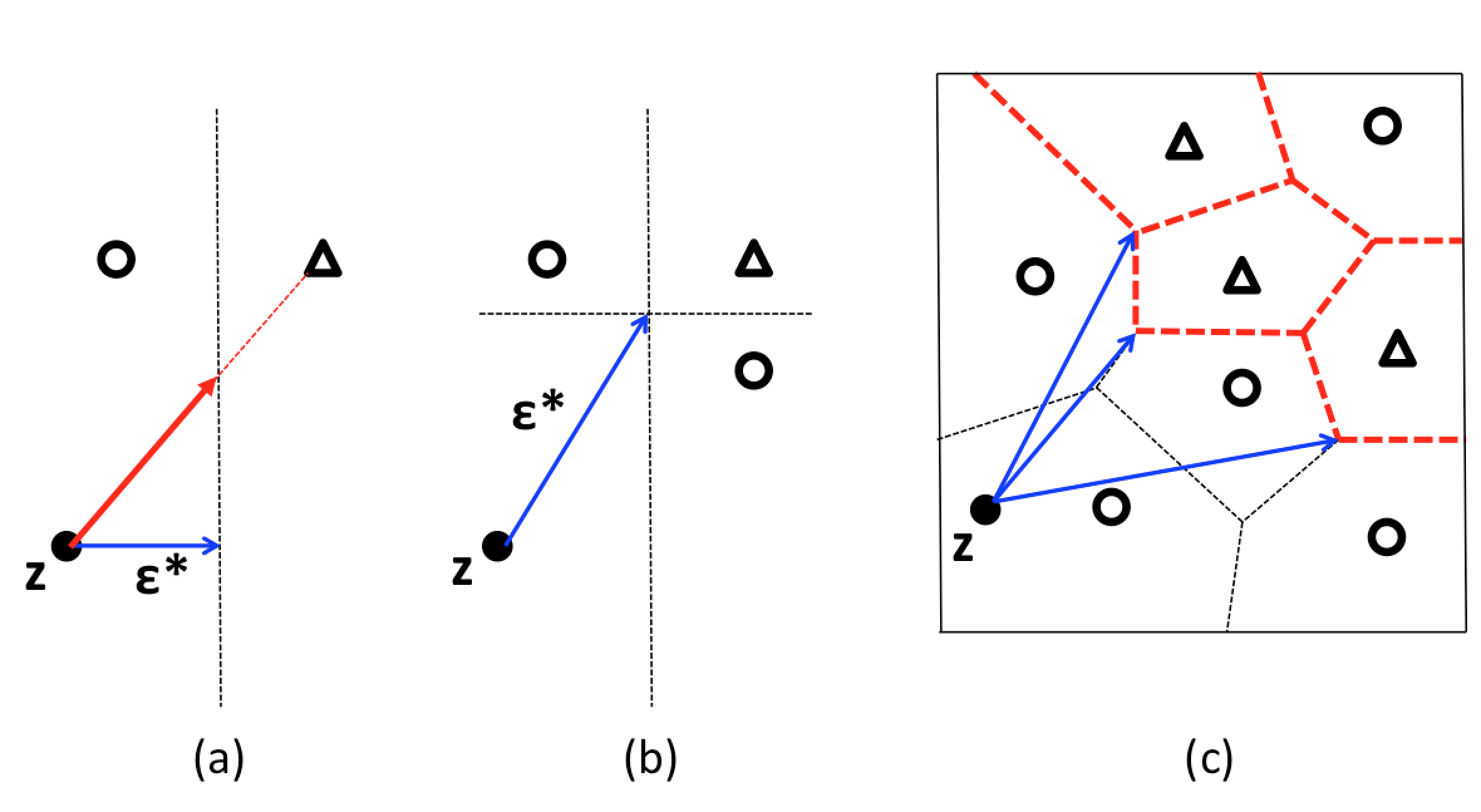}
    \caption{Illustration of the minimum adversarial perturbation for 1-NN model. The goal is to perturb $\vz$ to be classified as triangle. In (a), the red curve is the perturbation computed by \cite{sitwarin2019robustness} while the optimal solution (blue perturbation, $\epsilon^*$) could be much better, and the ratio can be arbitrary large by changing the angle between red and blue.  (b) shows that projection to the bisection hyperplanes may not be optimal; one also needs to consider intersections of several bisections which can be exponentially many. (c) shows that the optimal perturbation can be computed by evaluating the distance from $\vz$ to each Voronoi cell of  triangle instances.  
    }
    \label{fig:illustration}
\end{figure}

At first glance, the minimum adversarial perturbation seems to be easy to compute for the $1$-NN model. For instance, \cite{sitwarin2019robustness} mentioned that the minimum adversarial perturbation has to be on the straight line connecting $\vz$ and one of the training instances belonging to a different class ($y_i\neq 1$), so a simple linear time algorithm can solve this problem. Unfortunately, this claim is not true. In Figure \ref{fig:illustration}, we show that the optimal perturbation may not be on the lines connecting two points and furthermore, only checking the line segments can find an arbitrary bad solution.
Other previous approaches try to form a continuous approximation of NN classifiers~\cite{sitwarin2019robustness,PapernotMG16,dubey2019defense}, and clearly, they cannot find the optimal perturbation. 

\paragraph{Connection to Voronoi diagrams and a solution for low-dimensional cases.}

In fact, the decision boundary of a 1-NN model can be captured by the Voronoi diagram (see Figure~\ref{fig:illustration}(c)). In the Voronoi diagram, each training instance $\vx_i$ forms a cell, and the decision boundary of the cell is captured by the convex boundary formed by bisections between $\vx_i$ and its neighbors. One can thus obtain the minimum adversarial perturbation by computing the distances from $\vz$ to all the cells with $y_i\neq 1$. However, to compute the distance, we need to check all the faces (captured by one bisection hyperplane) and angles (intersections of more than one bisection hyperplanes) of the cell. 
   
    For 2-dimensional space ($d=2$), 
    it has been shown in \cite{aurenhammer2000voronoi} that each cell can only have finite faces and angles and there exists a polynomial time algorithm for computing a Voronoi diagram.
    In general, for $d$-dimensional problems with $n$ points,
    Voronoi diagram computation requires $O(n\log n + n^{\ceil{\frac{d}{2}}})$ time, which works for low-dimensional problems. 
    However, time complexity grows exponentially with dimension $d$,
    so in general, it is hard to use this algorithm unless $d$ is very small. 

\section{Primal-dual quadratic programming formulation}

Usually, finding the minimum adversarial perturbation is hard. Computing minimum adversarial perturbations for ReLU networks and tree ensembles are both NP-hard~\cite{katz2017reluplex,kantchelian2016evasion}. Also, as discussed in the previous section, we can connect it to Voronoi diagram computation, but the solver will require exponential time in dimensionality. So is it NP-hard to compute the minimum adversarial perturbation for $1$-NN? 
Surprisingly, it is not as we will demonstrate below. 

\subsection{Quadratic (primal) problems for minimum adversarial perturbation}

We consider the 1-NN model. For a given instance $\vz$, if we want to perturb it so that 
$\vz + \vdelta$ is closer to $\vx_j$ with $y_j\neq 1$ than to all class-$1$ instances,
then the problem of finding the minimum perturbation can be formulated as: 
\begin{equation}
\label{eq:oo}
 \epsilon^{(j)} = \min_{\vdelta} \ \frac{1}{2} \vdelta^T \vdelta \
\text{ s.t. }  \|\vz+\vdelta - \vx_j\|^2 \leq \|\vz+\vdelta - \vx_i\|^2, \ \ \forall i,  y_i=1. 
\end{equation}
Each constraint can be rewritten as
$\vdelta^T (\vx_j-\vx_i) + \frac{\|\vz-\vx_i\|^2 - \|\vz-\vx_j\|^2}{2} \geq 0$.
Therefore \eqref{eq:oo} becomes
\begin{align}
\label{eq:primal}
    \epsilon^{(j)} = \min_{\vdelta: A\vdelta + \vb \geq 0} \ & \ \{\frac{1}{2} \vdelta^T \vdelta \}:=P^{(j)}(\vdelta),
    \end{align}
    where $A\in \R^{n\times d}$ and $\vb\in \R^{n}$, for each row $i$ with $y_i = 1$, $\va_i=(\vx_j-\vx_i )$ and $b_i=\frac{\|\vz-\vx_i\|^2 - \|\vz-\vx_j\|^2}{2}$ respectively  ($\bm{0}$ otherwise). 
    By solving the quadratic programming (QP) problem~\eqref{eq:primal} for each $\{j: y_j\neq 1\}$, the final minimum adversarial perturbation norm is 
        $\epsilon^* =  \min_{j: y_j\neq 1} \sqrt{2\epsilon^{(j)}}$. 
        It has been shown that convex quadratic programming can be solved in polynomial time~\cite{kozlov1980polynomial}, so our formulation leads to a {\bf polynomial time algorithm} for finding $\epsilon^*$.  
         
\subsection{Dual quadratic programming problems}

We also introduce the dual form of each QP, which is more efficient to solve in practice and will lead to a verification perspective of evaluating adversarial robustness. The dual problem of \eqref{eq:primal} can be written as
\begin{equation}
    \max_{\vlambda \geq 0} \{- \frac{1}{2} \vlambda^T A A^T \vlambda - \vlambda^T \vb\}:=D^{(j)}(\vlambda), 
    \label{eq:dual}
\end{equation}
where $\vlambda \in \R^{n}$ are the corresponding dual variables. 
The derivation is easy, but for completeness, we include it in Appendix~\ref{app:primal_dual}. 
The primal-dual relationship connects primal and dual variables: 
\begin{equation*}
    \vdelta = A^T \vlambda.
\end{equation*}
Based on the weak duality, we have  
    $D^{(j)}(\vlambda) \leq P^{(j)}(\vdelta)$ for any dual feasible solution $\vlambda$ and primal feasible solution $\vdelta$. 
    Furthermore, based on Slater's condition we can easily show that strong duality holds ($D^{(j)}(\vlambda^*) = P^{(j)}(\vdelta^*)$) if 
    $\vx_j\neq \vx_i$, $\forall i, y_i = 1$.
    \footnote{One can observe that if the condition holds ($y_j\neq 1$) then a small ball around $\vdelta = \vx_j - \vz$ will be feasible solutions which satisfies Slater's condition, implying strong convexity.} 
    Based on strong duality, we have
    \begin{equation}
    \frac{1}{2} (\epsilon^*)^2 \!=\! \min_{j:y_j\neq 1} \{ P^{(j)}(\vdelta^*)\} \!=\! \min_{j:y_j\neq 1} \{ \max_{\vlambda\geq 0} D^{(j)}(\vlambda) \} \!\geq\!
    \min_{j:y_j\neq 1} \{  D^{(j)}(\vlambda^{(j)}) \} \text{ with feasible $\vlambda^{(j)}$}, 
    \label{eq:verification}
    \end{equation}
        so any feasible solution $\{\vlambda^{(j)}\}$ leads to a lower bound of the minimum adversarial perturbation.
        In summary, we conclude the primal-dual relationship between 1-NN attack and verification: 
    \begin{compactitem}
        \item A primal feasible solution of $P^{(j)}$ for any $y_j \neq 1$ is a successful attack and gives us an upper bound of $\epsilon^*$. Therefore, one can solve QPs with a subset of $j$; usually a $\vx_j$ closer to $\vz$ will lead to a smaller adversarial perturbation, so in practice we can sort $\vx_j$ by the distance to $\vz$, solve the subproblems one by one, and stop at any time. It will give a valid adversarial perturbation. After solving all the subproblems, the result will be $\epsilon^*$. 
        \item A set of dual feasible solutions $\{\vlambda^{(j)}\}_{j: y_j\neq 1}$ will give a lower bound of $\epsilon^*$ according to \eqref{eq:verification}. So any heuristic method for setting up a set of dual feasible solutions will give us a lower bound which can be used for robustness verification. After solving all the dual problems exactly, we will get the tightest lower bound, which is also $\epsilon^*$. 
    \end{compactitem}
\paragraph{1-NN verification.}
Here we give an example of how to quickly set up dual variables to give a lower bound of the minimum adversarial perturbation without solving any problem. For a dual problem $D^{(j)}$, consider only having one variable $\lambda^{(j)}_i$ to be nonzero while fixing all the rest variables zero, the optimal closed-form solution will be 
\begin{equation}
    \lambda^{(j)}_i = \max(0, -\frac{b_i}{\|\va_i\|^2}), \ \  D^{(j)}([0, \dots, 0, \lambda^{(j)}_i, 0, \dots, 0]) =
    \frac{\max(-b_i, 0)^2}{2\|\va_i\|^2}.
    \label{eq:bound_cd}
\end{equation}
Note that $b_i=\frac{\|\vz-\vx_i\|^2 - \|\vz-\vx_j\|^2}{2}$
and $\|\va_i\|^2 = \|\vx_j-\vx_i\|^2$ both can be computed easily, so a guaranteed lower bound of $\epsilon^*$ can be computed easily: 
\begin{equation}
    \underline{\epsilon} = \min_{j: y_j\neq 1} \big( \max_{i: y_i=1} \frac{\max(\|\vz-\vx_j\|^2-\|\vz-\vx_i\|^2, 0)}{2\|\vx_j-\vx_i\|}\big) \leq \epsilon^*. 
    \label{eq:lower_bound}
\end{equation}
This value has an interesting geometrical meaning. See Appendix~\ref{app:geometric} for more details.  
In general, we can also get improved lower bounds by solving more coordinates for each subproblem. 

\subsection{Solving the QP problems efficiently}
    
    Now we discuss how to efficiently solve a series of QP problems $\{D^{(j)}\}_{j: y_j\neq 1}$ in practice. Although we can do this in polynomial time, in practice a naive algorithm is still too slow. Note that we have totally $O(n)$ quadratic problems and each QP has $O(n)$ variables to solve, so roughly more than $O(n^3)$ time complexity is required for doing this naively. Calling a commercial quadratic programming solver will take $20$ seconds for $n=6000$ when solving only one QP problem. 
    In the following, we show how to
    solve ``all'' QP problems in $3$ seconds.
    
    First, we find that a greedy coordinate ascent algorithm can be efficiently applied to solve the dual QP problem~\eqref{eq:dual}. This is mainly due to the sparsity of the solution---if $\vlambda^*$ is dual optimal, then a nonzero $\lambda^*_i$ means the primal constraint $\|\vz + \vdelta - \vx_j\|^2 = \|\vz + \vdelta - \vx_i\|^2$, so the optimal $\vz+\vdelta$ will be on the bisection of $(\vx_i, \vx_j)$. Therefore, if $\|\vlambda^*\|_0=q$  then the optimal solution is the intersection of $q$ bisection hyperplanes, which means $q$ is usually small.
    For instance, on {MNIST} dataset when we test on 100 subproblems, the average number of $\|\vlambda^*\|_0$ is only $50.06$, with $7612.59$ dual variables per subproblem.
    The sparsity of the solution motivates the use of the greedy coordinate ascent algorithm. Starting from $\vlambda=\vzero$, we maintain the gradient vector with $\vg = -AA^T\vlambda-\vb$ and every time we pick the variable with the largest projected gradient
    \begin{equation*}
        i^* = \arg\max_i |( \max(\vlambda + \vg, 0)-\vlambda)_i|
    \end{equation*}
    and then update a single variable $\lambda_{i^*}\leftarrow \max(\lambda_{i^*} + g_{i^*}/\|\va_{i^*}\|^2, 0)$. This is similar to the SMO method proposed for training kernel SVM~\cite{platt1998sequential,chang2011libsvm}, but since there is no equality constraint we only need to pick one variable at a time. Since there are only a few nonzero $\lambda$s, the algorithm usually converges much quicker than standard quadratic optimization solvers. 
    
    Second, we propose a screening rule to remove variables in each dual QP problem~\eqref{eq:dual}. There are only a few nonzero variables, and our screening rule will reduce the size of variables {\it before solving the problem}. We introduce the following lemma:
    \begin{lemma}
    \label{lm:screening_1}
For a specific quadratic problem $P^{(j)}(\vdelta)$, the optimal dual solution has $\lambda^*_i=0$ if 
\begin{equation}
    -(\|\vz-\vx_i\|^2 -\|\vz-\vx_j\|^2)/2 + \|\vx_j-\vx_i\|\|\vdelta^*\| < 0, 
    \label{eq:screening_1}
\end{equation}
where $\vdelta^*$ is the optimal solution. 
\end{lemma}
The proof is in Appendix~\ref{app:screening_1}. 
Note that checking \eqref{eq:screening_1} does not need to solve the QP problem. 
To conduct the screening rule in \eqref{eq:screening_1}, we need to have an estimation of $\|\vdelta^*\|$ or its upper bound. A naive upper bound $\|\vdelta^*\|\leq \|\vx_j - \vz \|$ can be used for running the screening rule.  
    
With the methods mentioned above, each dual QP can be solved efficiently. However, there are $O(n)$ QPs in total, and solving all of them is still expensive. However, the final solution $\epsilon^*$ only depends on the minimum among solutions of all  QPs, so {\it can we remove most of the irrelevant QPs?} 

We can use primal-dual relationship for removing most of the QP problems before solving them. Assume we have a primal solution $\bar{\vdelta}$ then the minimum adversarial perturbation norm $ \epsilon^* \leq \|\bar{\vdelta}\|$,
so every dual problem with 
$D^{(j)}(\vlambda) > \bar{\vdelta}^T\bar{\vdelta}/2$
for some $\vlambda$ can be removed. For a subproblem with respect to $\vx_j$, based on \eqref{eq:bound_cd} we know the subproblem can be removed if
\begin{equation}
    \bar{\vdelta}^T \bar{\vdelta} < \max(-b_i, 0)^2/\|\va_i\|^2
    \label{eq:screening_subpb}
\end{equation}
for some $i$, thus we can use \eqref{eq:screening_subpb} to remove some unimportant subproblems. 
In practice, we sort the subproblems in ascending order of $\|\vz- \vx_j\|$ and iteratively run the screening rule after solving one more subproblem. As a result, most of the subproblems can be removed without solving them, and we achieve significant speedup. Our overall algorithm is illustrated in Algorithm~\ref{alg:qp}.

\begin{algorithm}[h]
\SetAlgoLined
\KwIn{Target instance $\vz$, database $\{(\vx_j, y_j)\}_{j=1}^n$.}
Initial $\epsilon= \infty$ \;
 Sort subproblems $\{D^{(j)}\}_{j: y_j\neq 1}$ by ascending distances of $\|\vz-\vx_j\|$\;
 \For{each $j$ (according to the sorted order)}{
    \If{not screenable via \eqref{eq:screening_subpb}}{
        Solve the subproblem via greedy coordinate ascent with screening rule \eqref{eq:screening_1}\;
        Update $\epsilon$ if we get a smaller value\;
    }
 }
 \caption{Computing minimum adversarial perturbation}
 \label{alg:qp}
\end{algorithm}

\subsection{Extending to $\ell_1$ and $\ell_\infty$ norms} \label{sec:lp}

Sometimes people are interested in finding the minimum $\ell_\infty$ or $\ell_1$ norm adversarial perturbation (replacing the  $\ell_2$ norm in \eqref{eq:attack}). Those can be solved similarly using our framework but will require linear programming instead of quadratic programming. For example, the minimum $\ell_\infty$-norm adversarial perturbation can be formulated as 
\begin{align*}
    \epsilon^{(j)} = \min_{\vdelta} \   v \ \ 
    \text{ s.t. } \  \ A\vdelta + \vb \geq  0, \ \ 
    v\geq \delta_i \geq -v  \ \ \forall i=1, \dots, d.
\end{align*}
A similar formulation can be done for the $\ell_1$ case. This can also be solved efficiently by linear programming solvers and the primal-dual relationship also holds.

\subsection{Extending beyond 1-NN}

We can extend our approach to $K$-NN with $K>1$ by adding more constraints.
Taking the $3$-NN model and binary classification as an example, we can list all the possible combinations of $\{(j_1, j_2, j_3)\mid y_{j_1}=y_{j_2}=2, y_{j_3}=1\}$ and then solve the following QP problem to force $\vz + \vdelta$ to be closer to $\vx_{j_1}, \vx_{j_2}$ than to all the class-$1$ instances except $\vx_{j_3}$:  
    \begin{align*}
 \epsilon^{(j_1, j_2, j_3)} \!=\! \min_{\vdelta} \    \frac{1}{2} \vdelta^T \vdelta  \ \ 
\text{ s.t. }   \ \|\vz+\vdelta - \vx_j\|^2 \leq \|\vz+\vdelta - \vx_i\|^2, \ \   
\forall i, i\neq j_3, y_i=1,
\ \ \ j \in \{j_1, j_2\}.  
\end{align*}
There will be $O(2n)$ constraints so will be more expensive to solve. For general $K>1$, we can write a similar formulation with $O(nK)$ constraints, but since the QP still has a sparse solution, greedy coordinate ascent can still solve a subproblem efficiently.  
Using this we can still obtain an upper and lower bound, corresponding to attack and verification. For an upper bound (attack), we just need to heuristically choose some $(j_1, j_2, j_3)$ tuples according to the distance to $\vz$ and solve some QPs (more details in Appendix~\ref{app:solve_bigK}). For a lower bound, we can use the similar formulation to \eqref{eq:lower_bound} as below:

\paragraph{Efficient verification for $K>1$.}
We can apply \eqref{eq:verification} efficiently even for large $K$. Taking the $K=3$ case as an example, let $C_{i,j} = \frac{\max(\|\vz-\vx_j\|^2-\|\vz-\vx_i\|^2, 0)}{2\|\vx_j-\vx_i\|}$ then with some simple derivation we can get the verification bound for $K=3$ case
\begin{equation*}
    \min_{(j_1, j_2, j_3): y_{j_1}=2, y_{j_2} = 2, y_{j_3}=1} \big(\max_{i\neq j_3}( \max(C_{i, j_1}, C_{i, j_2}))\big) \geq \min_{(j_1, j_2): y_{j_1}=2, y_{j_2}=2}
    \max(D_{j_1},D_{j_2}) 
\end{equation*}
where $D_{j} = \min_{t, y_t=1} ( \max_{i\neq t, y_i = 1} C_{i, j})$, which is the second largest value among $C_{i,j}$ for all $i$. Therefore, we just need to choose the second smallest of $D_j$ among $\{j: y_j=2\}$. 
Note that this can be generalized to a general $K$ case, where the verification lower bound becomes: 
\begin{equation}
     \underline{\epsilon} := \{\kthmin_{j: y_j=2}\ ( \kthmax_{i:y_i=1} \  C_{ij} \ )\} \leq \epsilon^*, \ \ k=(K+1)/2,
    \label{eq:verification_k}
\end{equation}
which can be computed efficiently with time complexity independent to $K$. 

%
%
%

\section{Experiments}

We show main results in Section~\ref{sec:main_results}, and analyze efficiency of our algorithm in Section~\ref{sec:efficiency}.
All experiments are run on a cloud server with one Intel E5-2650V4 CPU and one NVIDIA V100 GPU.
\subsection{Comparison of adversarial perturbations}
\label{sec:main_results}


We show that our formulation leads to better attack and verification algorithms.
Note that our QP framework leads to the following proposed algorithms for exact computation, verification, and attack:  \begin{compactitem}
    \item Exact: computes the exact minimum adversarial perturbation for 1-NN via Algorithm~\ref{alg:qp}.
    \item Verifier: computes a lower bound  for 1-NN via \eqref{eq:lower_bound} and for $K$-NN via \eqref{eq:verification_k}.
    \item QP-$1$ and QP-$10$: compute upper bounds (attack) for 1-NN via Algorithm~\ref{alg:qp} but only iterate over the top-$1$ and top-$10$ QP problems respectively.
    \item QP-greedy: computes an upper bound for $K$-NN by heuristically choosing QP subproblems (Appendix~\ref{app:solve_bigK}).
\end{compactitem}
Note that there is no existing algorithm for computing the exact minimum adversarial perturbation and no existing verification method for $K$-NN that can compute a lower bound. Therefore we are only able to compare with the following attack methods: 
\begin{compactitem}
    \item  Naive-$1$ and Naive-$10$~\cite{amsaleg2017vulnerability, sitwarin2019robustness}: compute upper bounds for 1-NN by moving towards a nearby \emph{other-class} instance (belonging to a class different from the one of the test instance). Naive-$10$ repeats the process for $10$ times and chooses the best perturbation. For $K>1$, Naive-1 moves towards a nearby size $(K+1)/2$ other-class cluster.
    \item Mean~\cite{sitwarin2019robustness}: computes an upper bound for $K$-NN by moving towards a class mean. The target class is chosen by the class mean distance to the test instance.
    \item Substitute~\cite{PapernotMG16}: computes an upper bound for 1-NN by attacking a smoothed variant of NN.
\end{compactitem}
All attack methods are tuned to have $100\%$ attack success rates, such that the perturbation is strictly the upper bound for the minimum adversarial perturbation.

\paragraph{Perturbations for 1-NN.} Experiments are performed on MNIST~\cite{lecun1998gradient} and Fashion-MNIST~\cite{xiao2017fashion}.
As Table~\ref{tab:1-nn-perturb} shows, Algorithm~\ref{alg:qp} (Exact) can efficiently compute the minimum adversarial perturbation.
Verifier can compute a reasonable lower bound without solving any QP problems exactly.
QP-$1$ and QP-$10$ are efficient and effective attack methods by iterating over only a few QP problems.

\begin{table}
\centering
  \caption{Mean perturbations and the total runtime of $100$ correctly classified test instances for 1-NN. 
  We repeat $5$ times to report the mean of total runtime. 
  }
  \begin{tabular}{rlcccc}
  \toprule
   & & \multicolumn{2}{c}{{MNIST}} & \multicolumn{2}{c}{{Fashion-MNIST}} \\
   & & Perturbation ($\ell_2$) & Runtime (s) & Perturbation ($\ell_2$) & Runtime (s) \\
    \midrule
   & Exact & $1.491$ & $177.507$ & $1.128$ & $130.795$ \\
   \midrule
   Lower bounds & Verifier & $1.370$ & $101.850$ & $1.073$ & $100.651$\\
   \midrule
   \multirow{5}{*}{Upper bounds} 
   & QP-$1$ & $1.530$ & $43.676$ & $1.142$ & $38.344$ \\
   & QP-$10$ & $1.491$ & $108.390$ & $1.128$ & $65.947$ \\
   & Naive-$1$ & $1.851$ & $145.299$ & $1.446$ & $115.820$  \\
   & Naive-$10$ & $1.755$ & $512.164$ & $1.413$ & $504.344$ \\
   & Mean & $4.561$ & $40.951$ & $4.179$ & $39.423$ \\
   & Substitute & $1.616$ & $164.153$ & $1.264$ & $145.797$\\
   \bottomrule
  \end{tabular}
  \label{tab:1-nn-perturb}
\end{table}

%
%



\paragraph{Perturbations with larger $K$.}
Experiments are performed on Binary-MNIST, where label $8$ and label $0$ are used. Verifier and QP-greedy compute tight lower bounds and upper bounds respectively as shown in Table~\ref{tab:knn}. More results are in Appendix~\ref{app:k-verification}.

\begin{table}
  \centering
  \caption{Mean perturbations of $100$ correctly classified test instances for $K$-NN.}
  \begin{tabular}{llccccc}
    \toprule
     & & $K=1$ & $K=3$ & $K=5$ & $K=7$ & $K=9$ \\
    \midrule
    Lower bounds& Verifier & $2.268$ & $2.230$ & $2.201$ & $2.193$ & $2.183$\\
    \midrule
    \multirow{3}{*}{Upper bounds}
    & QP-greedy & $2.494$ & $3.089$ & $3.417$ & $3.636$ & $3.786$\\
    & Naive-$1$ & $2.894$ & $3.718$ & $3.903$ & $4.085$ & $4.173$ \\
    & Mean & $5.282$ & $5.241$ & $5.205$ & $5.191$ & $5.176$ \\
    \bottomrule
  \end{tabular}
  \label{tab:knn}
\end{table}

\paragraph{Comparing robust error  under $\ell_{\infty}$-norm perturbation.}
In this experiment, we compare the robustness of 1-NN model with the CNN model on two datasets in Table~\ref{tab:my_label}. The CNN model has two convolutional layers and two fully connected layers with ReLU activations. 
We observe that neural nets have better test error, which is known in the literature. However,  if we compare the error under the same amount of attack, 1-NN outperforms neural nets on these two datasets. Furthermore, since 1-NN is easy to verify using our approach, and it is NP-hard to compute the minimum adversarial perturbation for a neural network, 1-NN can have much better verifiable robust error than neural nets. 
Note that we use one of the state-of-the-art verification methods in~\cite{kolter2017provable} for computing the verifiable robust error of neural nets. 
We do not claim 1-NN is a better model than neural nets. For more complex datasets such as CIFAR or ImageNet, the 1-NN method will lead to bad clean error, so it is not comparable with neural nets. However, we think this experiment suggests that for some simple data, NN models could be a better choice in terms of the robustness error. 

\begin{table}
    \centering
    \caption{Comparison of robustness for 1-NN and neural networks with the same $\ell_\infty$ perturbation.}
    \begin{tabular}{cccccc}
    \toprule
    Dataset     & Model  & $\epsilon$ & Test error & Attack error & Verifiable robust error \\
    \midrule
    \multirow{2}{*}{MNIST} & Neural Net & 0.1 & {\bf 1.07\%} & 81.68\% & 100\% \\
     & 1-NN & 0.1 & 3.41\% & {\bf 27.06\%} & {\bf 27.06\%} \\
    \midrule
    \multirow{2}{*}{Fashion-MNIST} &  Neural Net  & 0.1 & {\bf 9.36\%} & 81.85\% & 100\% \\
     & 1-NN &  0.1 & 23.47\% & {\bf 78.25\%} & {\bf 78.25\%} \\
    \bottomrule
    \end{tabular}
    \label{tab:my_label}
\end{table}

\subsection{Efficiency of our algorithm}\label{sec:efficiency}
\begin{figure}
\centering
\begin{minipage}{.5\textwidth}
  \centering
  \includegraphics[height=.6\linewidth]{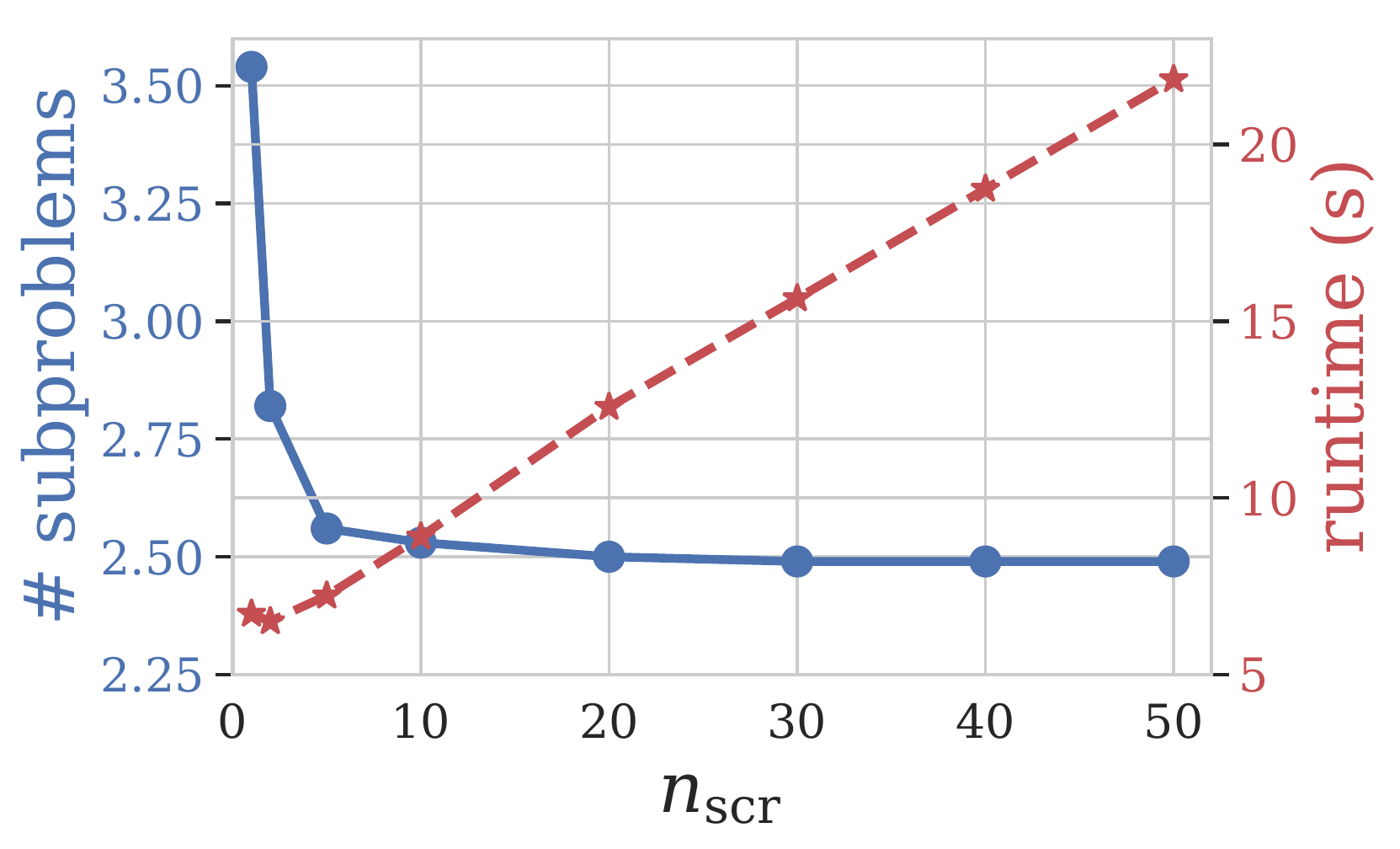}
  \captionsetup{width=1\linewidth}
  \captionof{figure}{Screening trade-off.}
  \label{fig:tradeoff}
\end{minipage}%
\hfill
\begin{minipage}{.5\textwidth}
  \centering
  \includegraphics[height=.6\linewidth]{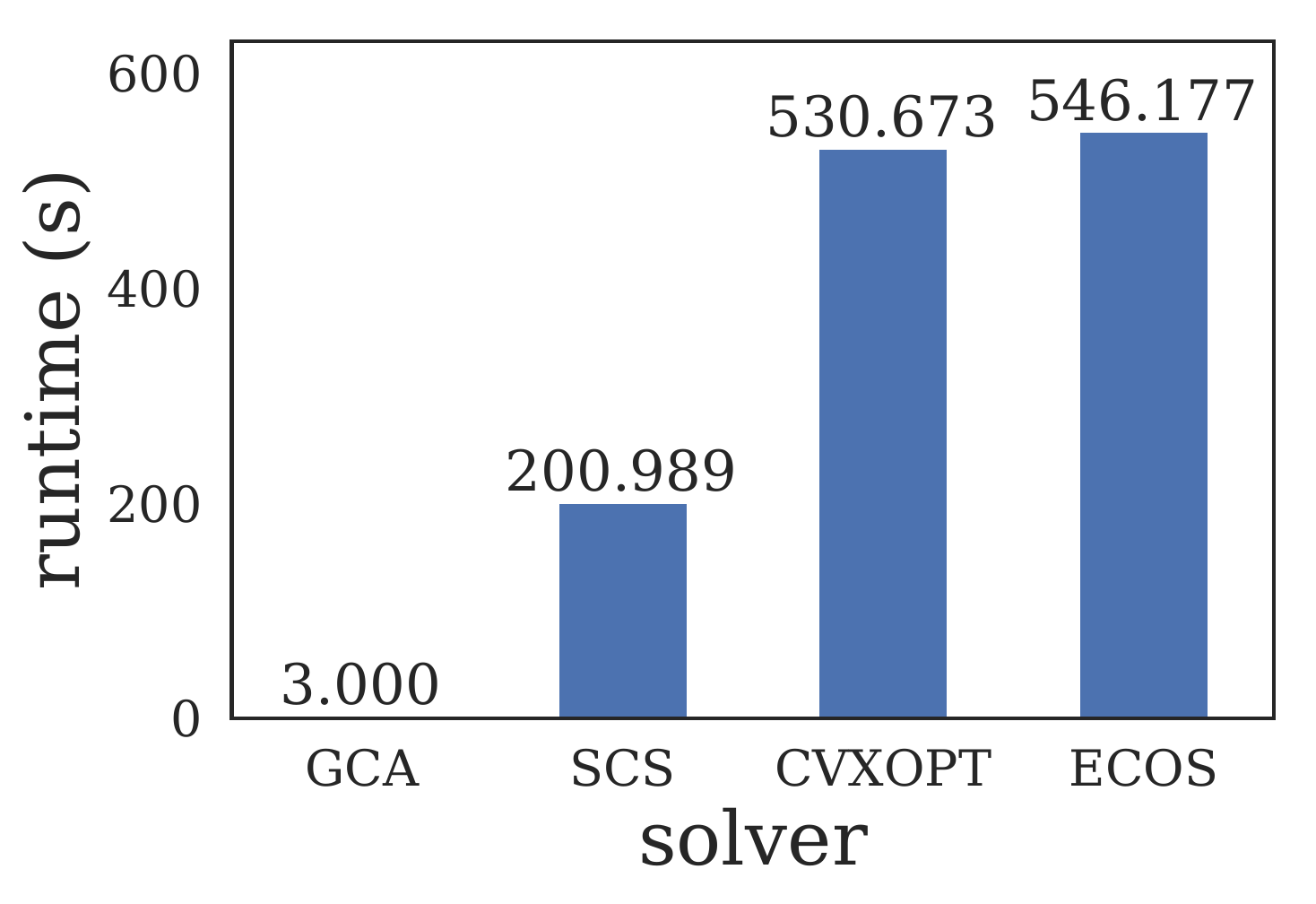}
  \captionsetup{width=1\linewidth}
  \captionof{figure}{Runtime of QP solvers.}
  \label{fig:solver-efficiency}
\end{minipage}
\end{figure}

We already show Algorithm~\ref{alg:qp} is efficient in Table 1. It has three components: sorting, screening rules for reducing the number of QPs, and the greedy coordinate ascent solver for each QP. We leave the experiments about sorting in Appendix~\ref{app:sort} and talk about the other two in detail. 
\paragraph{Screening. }
In the MNIST case, for every test instance, 
we have to solve about $54,000$ (all other-class instances) QP problems without screening.
While with sorting and screening, only $2.53$ QP problems on average are left to solve. 
Therefore screening improves efficiency significantly.
The screening parameter $n_\text{scr}$ (number of $i$s chosen for each $j$ in~\eqref{eq:screening_subpb}) is also an important parameter for efficiency.
There is a trade-off between the number of screened subproblems and the screening overheads controlled by $n_\text{scr}$.
We plot the tradeoff on MNIST in Figure~\ref{fig:tradeoff}. This shows a very small $n_\text{scr}$ is enough, and we choose 8 for our experiments. 



\paragraph{Greedy coordinate ascent.}
Due to the sparsity of each QP problem, the greedy coordinate ascent solver is much more efficient than other standard QP solvers. 
To verify this, we compare  greedy coordinate ascent  with
SCS~\cite{donoghue2016conic}, CVXOPT and ECOS~\cite{domahidi2013ecos} for solving these QP problems. 
We fix everything the same (with the same screening rule and sorting technique) while only change the QP solver. 
Since it is difficult for standard QP solvers to deal with high dimensional problems, $1/10$ training samples of {MNIST} are used.
The results are presented in Figure~\ref{fig:solver-efficiency}. Greedy coordinate ascent is faster than other solvers by more than 60 times for computing $K$-NN robustness. 



\section{Conclusion}
In this paper, we show that computing the minimum adversarial perturbation of $K$-NN models can be formulated as a series of quadratic programming problems. This framework is the first algorithm that can compute the minimum adversarial perturbation, and we propose an efficient solver such that the computation time is comparable with and often faster than previous attack algorithms. Furthermore, our framework also motivates the first algorithm for verifying $K$-NN robustness from the dual aspect. 

%
\medskip
\bibliographystyle{plain}
\bibliography{ref}

\newpage
\appendix
\section{Appendixes}
\subsection{Derivation of the dual form}
\label{app:primal_dual}
Consider the primal problem in~\eqref{eq:primal}. The Lagrangian can be written as
\begin{equation*}
    L(\vdelta, \vlambda) = \frac{1}{2} \vdelta^T \vdelta - \vlambda^T (A\vdelta+b) 
\end{equation*}
The dual problem is then
\begin{align*}
    \max_{\vlambda \geq 0} \min_{\vdelta} L(\vdelta, \vlambda)  
\end{align*}
Taking derivative of Lagrangian we get
\begin{equation*}
    \frac{\partial }{\partial \vdelta}L =
    \vdelta - A^T \vlambda=0, 
\end{equation*}
which gives us the primal-dual relationship $\vdelta = A^T \vlambda$. 
Substitute this back to the dual problem we get
\begin{equation*}
    \max_{\vlambda \geq 0 } - \frac{1}{2} \vlambda^T A A^T \vlambda - \vlambda^T \vb. 
\end{equation*}

\subsection{Geometric meaning of our verification bound}
\label{app:geometric}
Here we discuss the geometric meaning of the following verification bound for $1$-NN (derived in~\eqref{eq:lower_bound}): 
\begin{equation}
    \underline{\epsilon} = \min_{j: y_j\neq 1} \big( \max_{i: y_i=1} \frac{\max(\|\vz-\vx_j\|^2-\|\vz-\vx_i\|^2, 0)}{2\|\vx_j-\vx_i\|}\big) \leq \epsilon^*. 
\end{equation}
The inner value is the distance between $\vz$ to the bisection between $\vx_i$ and $\vx_j$, which means if we want to perturb $\vz$ to make it closer to $\vx_j$ than $\vx_i$, the perturbation must be larger than the inner value. Then, if we want to perturb $\vz$ such that the nearest neighbor is $\vx_j$, we need to by-pass all the bisections so we need to take the $\max$ operation among all the distances to bisections. And a lower bound of $\epsilon^*$ can be computed by taking minimum over all the $\vx_j$.

\subsection{Proof of Lemma~\ref{lm:screening_1}}
\label{app:screening_1}

\begin{proof}
By definition we have $\nabla D^{(j)}(\vlambda) = - AA^T \vlambda - \vb$, so
\begin{align*}
    \nabla D^{(j)}(\vlambda^*) 
    &= -AA^T\vlambda^* - \vb\\
    &= -A\vdelta^* - \vb.
\end{align*}
\begin{align*}
    \nabla D_i^{(j)}(\vlambda^*) 
    &= -\va_i^T \vdelta^* - b_i \\
    &= (\vx_i-\vx_j)^T \vdelta^* -  \frac{\|\vz-\vx_i\|^2 - \|\vz-\vx_j\|^2}{2} \\
    &\leq \lVert \vx_i-\vx_j\rVert \lVert \vdelta^* \rVert -  \frac{\|\vz-\vx_i\|^2 - \|\vz-\vx_j\|^2}{2}.
    \label{eq:cauchy}
\end{align*}
Therefore, when \eqref{eq:screening_1} holds, by KKT conditions of the dual problem we know $\lambda_i^* = 0$.

\end{proof}

\subsection{Attack for the $K>1$ case} 
\label{app:solve_bigK}

Note that the problem is equivalent to forming $A = [A_1, A_2]$ where the $i$-th row of $A_1$ is $(A_1)_i=(\vx_{j_1}-\vx_i)$ and $(A_2)_i=(\vx_{j_2}-\vx_i)$ for all class-$1$ instances $\{i: y_i=1\}$. 
We can first choose $j_1$ to be the class-$2$ instance closest to $\vz$, then try different $j_2$ (sorting according to the distance to $j_1$). After solving each pair of $j_1, j_2$, we can try to remove one row of $A_1$ and $A_2$ which corresponds to $j_3$. Note that only removing $j_3$ with nonzero $\lambda_{j_3}$ can change the result and there are only few nonzero $\lambda$s, so we could simply try all of them. 

A greedy and more efficient version is illustrated in Algorithm~\ref{alg:qp-greedy}.

\begin{algorithm}[H]
\SetAlgoLined
\KwIn{Target instance $\vz$, database $\{(\vx_j, y_j)\}_{j=1}^n$, neighbor parameter $K$.}
\While{True}{
Select a size $\lceil(K+1)/2\rceil$ subset of other-class instances $S^-$ all with the same label\;
Solve a QP problem to make $S^-$ nearest to $\vz + \vdelta$\;
\If{feasible}{
break\;
}
}
Select a size $\lfloor (K-1)/2 \rfloor$ subset of \emph{same-class} instances $S^+$ (belonging to the class the same with the one of the test instance), of which dual variables are not $0$\;
Solve a QP problem to make $S^-$ nearest to $\vz + \vdelta$ without constraints on $S^+$\;
\Return $\vdelta$\;

 \caption{QP-greedy}
 \label{alg:qp-greedy}
\end{algorithm}

\subsection{More experimental results for $K$-NN verification}\label{app:k-verification}

\begin{figure}
\centering
\begin{minipage}{.5\textwidth}
  \centering
  \includegraphics[height=.6\linewidth]{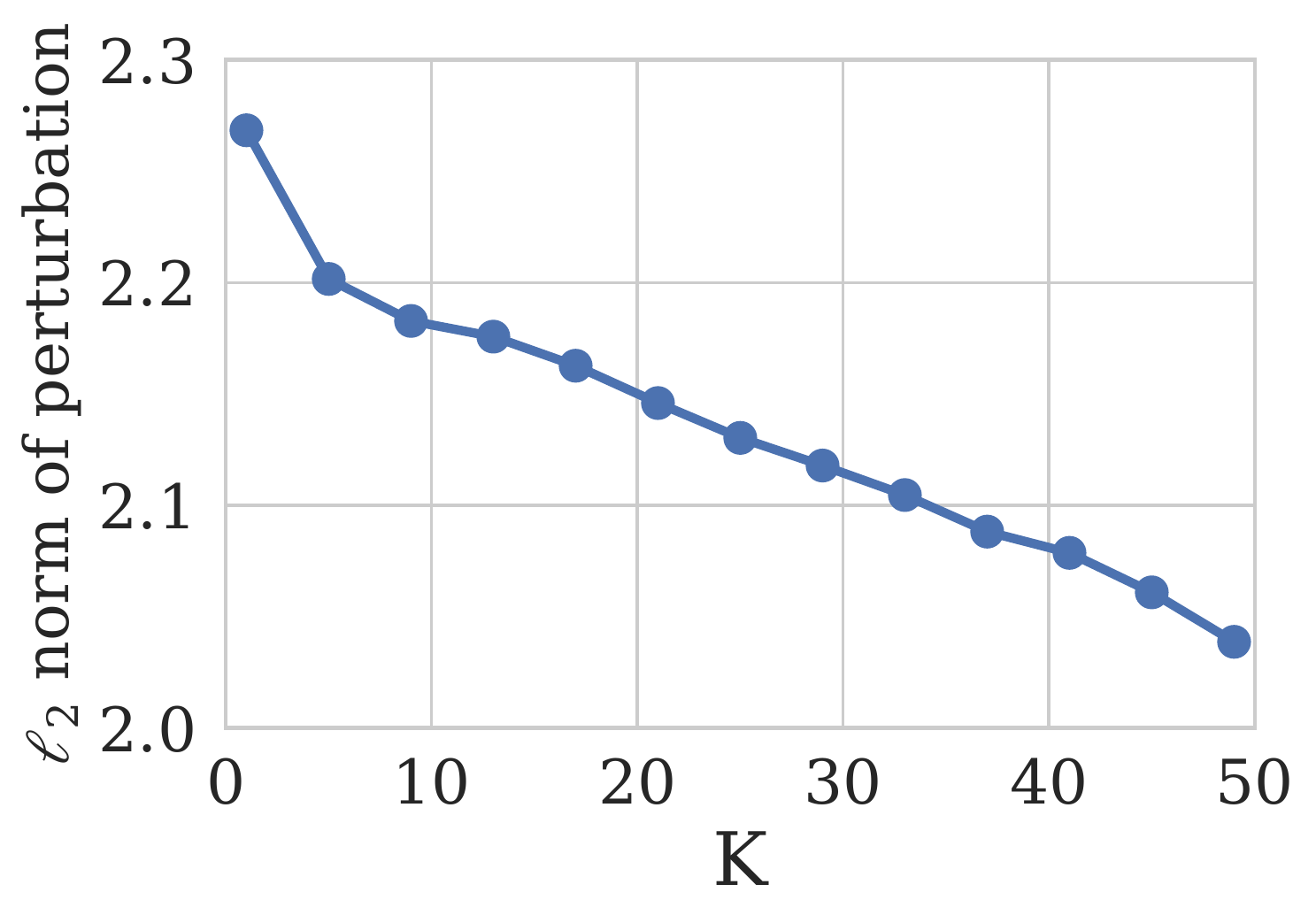}
  \captionsetup{width=.8\linewidth}
  \captionof{figure}{Verification (Lower bounds) via \eqref{eq:verification_k} for Binary-MNIST.}
  \label{fig:verification-binary}
\end{minipage}%
\hfill
\begin{minipage}{.5\textwidth}
  \centering
  \includegraphics[height=.6\linewidth]{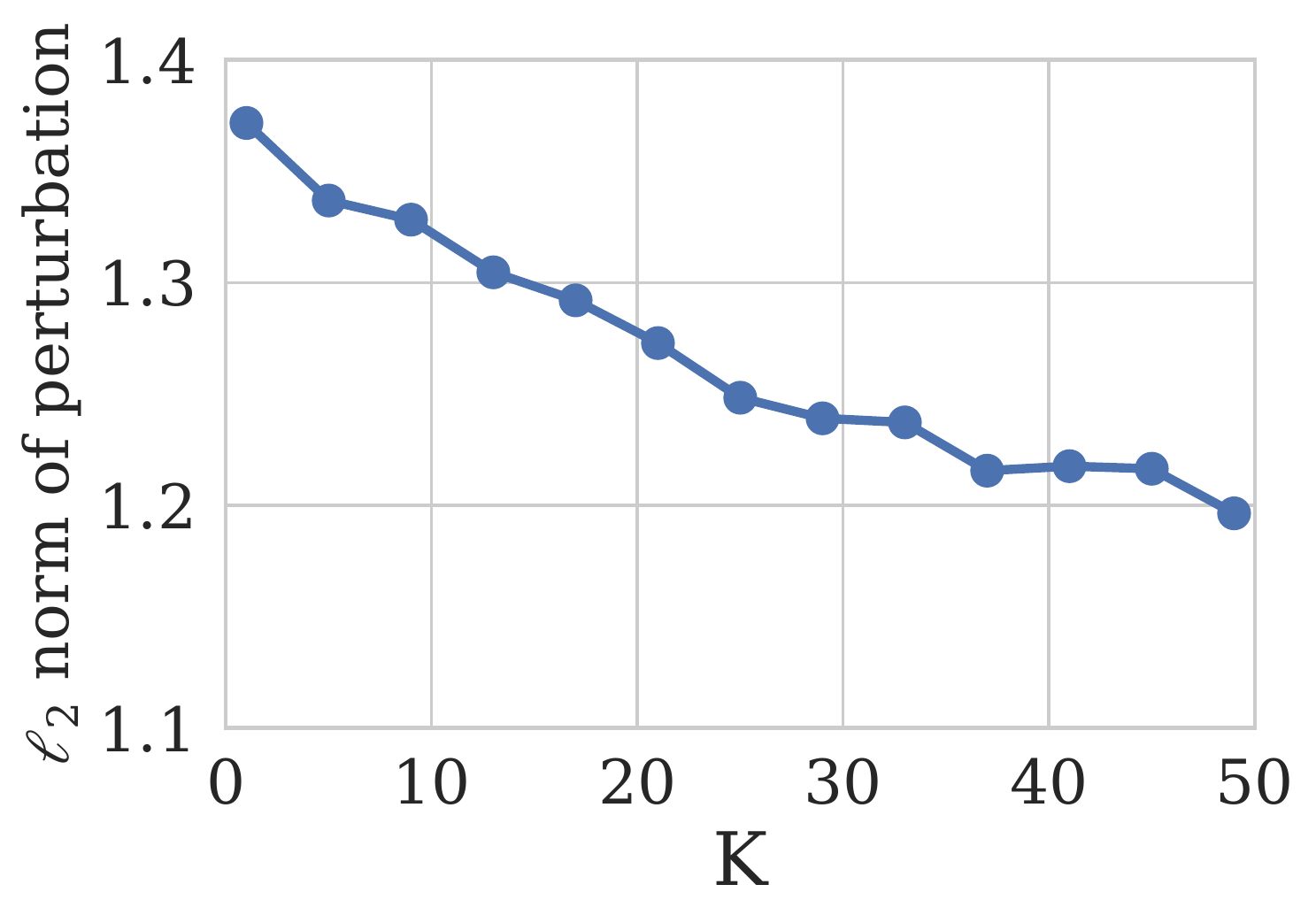}
  \captionsetup{width=.8\linewidth}
  \captionof{figure}{Verification (Lower bounds) via \eqref{eq:verification_k} for MNIST.}
  \label{fig:verification-multi}
\end{minipage}
\end{figure}

Since time complexity of our verification method for $K$-NN is independent to $K$,
we can efficiently compute lower bounds of the minimum adversarial perturbation for a large $K$.
Experimental results on Binary-MNIST are illustrated in Figure~\ref{fig:verification-binary}.

The verification method can be extended to the multi-class case.
A simple way is just taking the true label of the test instance as positive (label $1$), and the others as negative (label $2$).
It could be easily verified that $\underline{\epsilon}$ in \eqref{eq:verification_k} is still a lower bound. Experimental results on MNIST are illustrated on Figure~\ref{fig:verification-multi}.


\subsection{Effect of sorting in our Algorithm~\ref{alg:qp}} \label{app:sort}

We study whether sorting improves efficiency.
All training data of MNIST are used as training instances. $100$ correctly classified test instances are sampled randomly. All components of Algorithm~\ref{alg:qp} are employed, and $n_\text{scr} = 8$, i.e., $8$ positive instances are used for screening. We report the mean number of subproblems and the mean runtime of the $100$ test instances. No extra parallel mechanism is employed for test instances.
As Table~\ref{tab:sort} shows, sorting reduces the number of subproblems and improves efficiency.

\begin{table}
\centering
  \caption{The mean number of subproblems and the mean runtime of 100 correctly classified test instances. The only difference of the two rows is whether sorting negative examples as a first stage. No parallel mechanism is employed across test instances.}
  \begin{tabular}{lcc}
    \toprule
    & \# subproblems & runtime (s) \\
    \midrule
    w/ sorting & $2.530$ & $8.301$ \\
    w/o sorting & $25.490$ & $36.737$\\
    \bottomrule
  \end{tabular}
  \label{tab:sort}
\end{table}

\end{document}